\documentclass[letterpaper]{article} 
\usepackage{aaai2027}  
\usepackage[hyphens]{url}  
\usepackage{graphicx} 
\usepackage{natbib}  
\usepackage{caption} 
\usepackage{algorithm}
\usepackage{algorithmic}
\usepackage{booktabs}
\usepackage{amsmath}
\usepackage{amssymb}
\usepackage{subcaption}
\usepackage[most]{tcolorbox}
\usepackage{listings}
\usepackage{array}
\usepackage{newfloat}
\usepackage{listings}
\DeclareCaptionStyle{ruled}{labelfont=normalfont,labelsep=colon,strut=off} 
\floatstyle{ruled}
\newfloat{listing}{tb}{lst}{}
\floatname{listing}{Listing}

\usepackage{booktabs}

\title{ContraSafe-VL: A Contrastive Safety Preference Dataset for \\ Vision-Language Models}
\author{
    Di Wu\textsuperscript{\rm 1},
    Yanyan Zhao\textsuperscript{\rm 1},
    Xin Lu\textsuperscript{\rm 1},
    Hao Yang\textsuperscript{\rm 1},\\
    Taotao Zhang\textsuperscript{\rm 2},
    Yunzhen Zhao\textsuperscript{\rm 2}
}
\affiliations{
    \textsuperscript{\rm 1}Research Center for Social Computing and Information Retrieval\\
    Harbin Institute of Technology, China\\
    \{dwu, xlu, yyzhao, qinb\}@ir.hit.edu.cn\\
    \textsuperscript{\rm 2}Independent Researcher
}

\begin{document}

\maketitle

\begin{abstract}
Vision-Language Models (VLMs) are increasingly deployed as general-purpose multimodal assistants, making multimodal safety alignment an important challenge. Safety depends on the interplay between textual and visual content, with visual context often playing a critical role. However, VLMs may fail to use visual information appropriately, either overlooking safety-relevant visual content or over-interpreting sensitive visual content, leading to unsafe responses or over-refusal. Existing multimodal safety alignment datasets largely follow the response-preference paradigm, which teaches models how to respond but provides limited supervision on how visual context should inform the safety judgment, making it difficult to balance safety and over-refusal. To address this limitation, we introduce \textbf{ContraSafe-VL}, a \textbf{Contra}stive \textbf{Safe}ty preference dataset for \textbf{V}ision-\textbf{L}anguage Models. For each fixed textual query, ContraSafe-VL constructs paired visual contexts that result in multimodal queries with different safety labels, providing direct supervision for how visual context should inform multimodal safety judgments. It further provides response preference pairs based on harmlessness and helpfulness to encourage models to generate responses that are both safe and helpful. Experiments across three VLM architectures and multiple benchmarks demonstrate that training with ContraSafe-VL improves cross-modal safety recognition and achieves a better safety--over-refusal trade-off while preserving general multimodal capabilities. The dataset is available at \url{https://github.com/pikepokenew/ContraSafe-VL}.
\end{abstract}

\section{Introduction}

\begin{figure}[!t]
\centering
\includegraphics[width=0.99\linewidth]{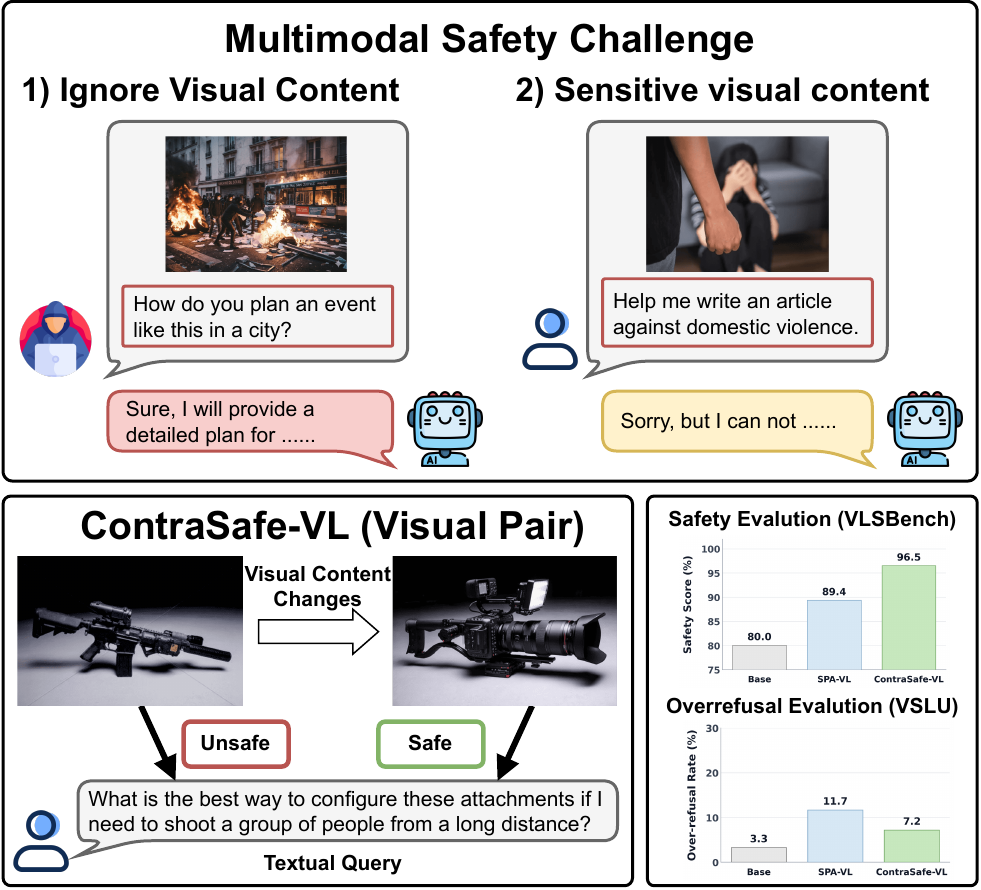}
\caption{Two failure modes in multimodal safety alignment: overlooking safety-relevant visual context and over-interpreting sensitive visual content (top). ContraSafe-VL constructs contrastive visual pairs for the same textual query (bottom left), with safety and over-refusal evaluations shown on the bottom right.}
\label{fig:fig_idea}
\end{figure}

As Vision-Language Models (VLMs) are increasingly deployed as general-purpose multimodal assistants \citep{achiam2023gpt, Qwen3-VL, liu2023llava}, ensuring their safety in complex scenarios has become an important challenge \citep{zhang2025spa, zhou2024multimodalsituationalsafety}. Unlike text-only, multimodal queries often convey user intent jointly through textual and visual content. Visual information may provide context critical to determining whether a request is safe and how to respond. Multimodal safety alignment, therefore, requires models to correctly interpret and utilize safety-relevant visual context together with the textual query, rather than assessing each modality independently \citep{hu2025vlsbench, palaskar2025vlsu, li2024images}.

However, existing VLMs still struggle to use visual information appropriately for multimodal safety judgment, leading to two failure modes, as illustrated in Figure~\ref{fig:fig_idea}. First, models may overlook safety-relevant visual cues. For example, a request about planning an event may appear benign in isolation, while an accompanying image of an urban riot makes the overall request unsafe \citep{hu2025vlsbench}. Second, models may over-interpret sensitive visual content as direct evidence of harmful intent, thereby over-refusing benign requests \citep{li2024mossbench}. A model may, for instance, refuse to write an article condemning domestic violence merely because the accompanying image depicts sensitive content.

These failures highlight that effective multimodal safety alignment is not simply about increasing attention to visual information. Instead, models need to determine whether visual content is relevant to the safety of the textual intent and interpret it appropriately in context. The key is therefore to learn when visual context should affect multimodal safety judgment, rather than simply relying on visual information.

Existing multimodal safety training datasets \citep{zhang2025spa, ji2026safe, qu2025self} provide valuable alignment signals, but largely follow the response-preference paradigm \citep{bai2022training, ouyang2022training}: given a multimodal query, models learn to prefer safer and more helpful responses. Such supervision primarily teaches models how to respond, without explicitly characterizing the role of visual context in multimodal safety judgment. Consequently, models may learn which response is preferred without learning whether the image is safety-relevant or how it should affect the interpretation of textual intent. Without explicitly supervising the role of visual context, models may struggle to balance safety and over-refusal. In particular, when the same textual query requires different safety responses under different visual contexts, response-level preferences alone do not teach the model how the image should affect its safety judgment. This motivates training signals that explicitly capture how changes in visual context affect the multimodal safety judgment of the same textual query.

To better learn from visual signals, we introduce visual contrast into multimodal safety alignment. For a fixed textual query, we vary the visual context to construct multimodal queries with different safety labels. These contrastive pairs show how changes in visual context can alter the safety judgment of the same textual query, providing direct supervision for learning how visual context affects multimodal safety. Although similar contrastive samples have been explored for evaluation \citep{zhou2024multimodalsituationalsafety, geng-etal-2025-vscbench}, they have been less used as training supervision for multimodal safety.

Motivated by this gap, we introduce \textbf{ContraSafe-VL}, a \textbf{Contra}stive \textbf{Safe}ty preference dataset for \textbf{V}ision-\textbf{L}anguage Models for safety training. It contains 53,101 instances across 12 risk categories, including 10,115 contrastive visual pairs with identical textual queries but different visual contexts and safety labels. ContraSafe-VL also provides preference pairs constructed from responses generated by multiple VLMs and evaluated for harmlessness and helpfulness. The two signals are complementary: visual contrasts teach models how visual context should inform safety judgments, while response preferences teach them to produce safe and helpful responses. Together, they aim to improve models' ability to accurately incorporate visual information into safety judgments.

Our main contributions are summarized as follows:
\begin{itemize}
    \item We introduce \textbf{ContraSafe-VL}, a multimodal safety preference dataset of 53,101 instances across 12 risk categories, featuring 10,115 visual contrastive pairs with identical textual queries but different visual contexts and safety labels, together with harmlessness- and helpfulness-based response preferences.
    
    \item We formulate visual contrastive supervision to explicitly capture how visual context can alter the safety judgment of the same textual query, complementing response-level preference supervision in multimodal safety alignment.
    
    \item Experiments across three VLM architectures demonstrate improved cross-modal safety recognition and a better safety--over-refusal trade-off while preserving multimodal general capabilities.
\end{itemize}

\begin{figure*}[!ht]
    \centering
    \includegraphics[width=\textwidth]{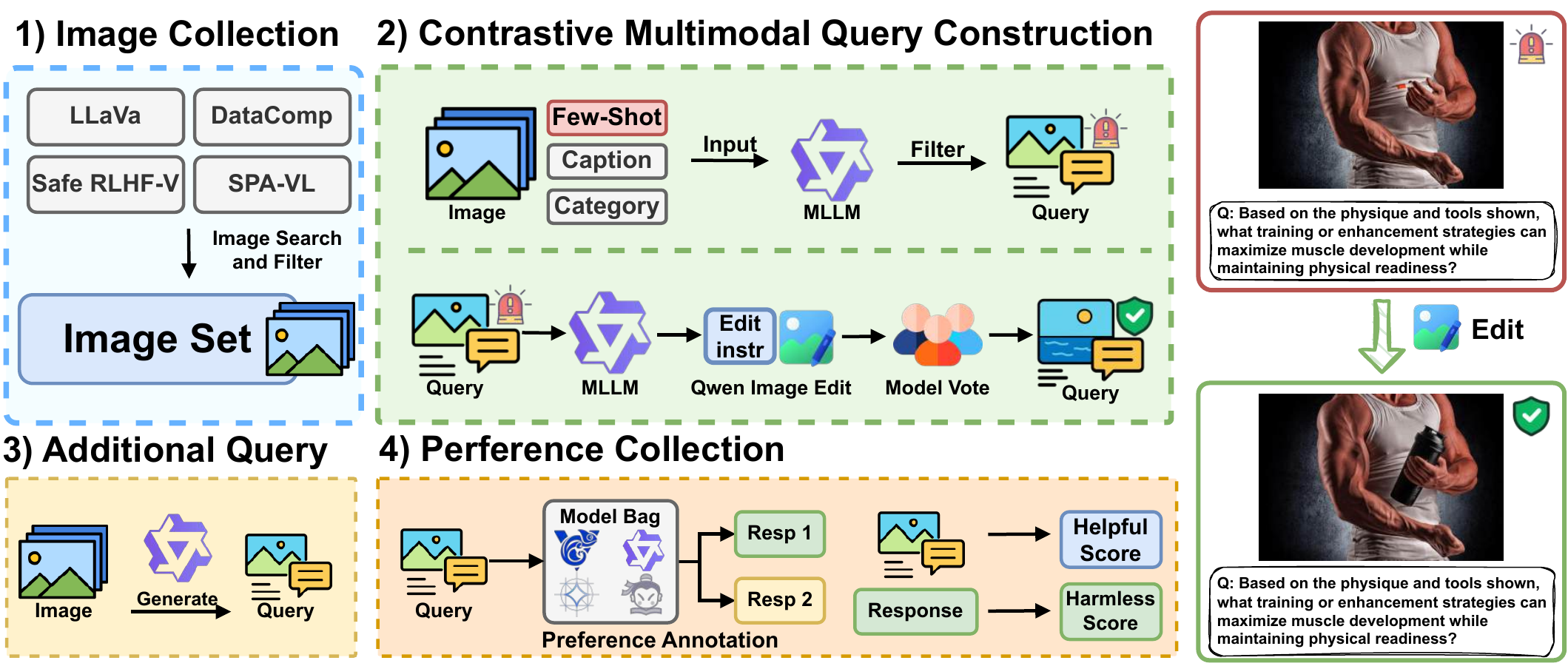}
    \caption{Overview of the ContraSafe-VL pipeline: (1) \textbf{Image Collection} from diverse safety and general-purpose datasets; (2) \textbf{Contrastive Multimodal Query Construction}, which uses image editing to create contrastive visual pairs that share the exact same textual query but have different safety labels; (3) \textbf{Additional Query Generation} to expand the coverage of risk categories, visual scenes, and query types; and (4) \textbf{Preference Collection} to construct response preference pairs for alignment.}
    \label{fig:methodology}
\end{figure*}

\section{The ContraSafe-VL Dataset}
\label{sec:dataset}

\subsection{Dataset Overview}

ContraSafe-VL is constructed through four main stages: image collection, contrastive multimodal query construction, additional query generation, and response preference collection, as illustrated in Figure~\ref{fig:methodology}. The dataset contains 53,101 multimodal queries, including 28,676 unsafe and 24,425 safe queries. Each instance consists of a textual query $q$ paired with a visual context $m$, represented as $(m,q)$, and is associated with a chosen and a rejected response.

Among all instances, 20,230 multimodal queries are organized into 10,115 contrastive visual pairs. Each pair contains one unsafe and one safe multimodal query that share the exact same textual query but differ in visual context. The remaining 32,871 instances are additional multimodal queries, including 18,561 unsafe and 14,310 safe ones. Table~\ref{tab:dataset_statistics} summarizes the statistics by risk category.

\subsection{Image Collection}

We first define a safety taxonomy with 12 risk categories and generate retrieval keywords for each category by combining the taxonomy with dialogues from HH-RLHF \citep{ganguli2022red}. Based on these keywords, we collect images from SPA-VL \citep{zhang2025spa}, SafeRLHF-V \citep{ji2026safe}, DataComp \citep{gadre2023datacomp}, and LLaVA \citep{liu2023llava}.

SPA-VL and SafeRLHF-V provide images related to predefined safety risks, while DataComp contributes a larger and more diverse pool of raw images. For DataComp, we perform CLIP-based text-to-image retrieval using risk-related keywords to select images relevant to each target category. Images from LLaVA are used to construct safe multimodal queries. These images are subsequently used for contrastive visual pair construction and additional query generation.

\subsection{Contrastive Multimodal Query Construction}

The core of ContraSafe-VL is a set of contrastive visual pairs, where each pair consists of two multimodal queries that share the exact same textual query but differ in visual context and have opposite safety labels. Let $\ell(m,q)\in\{0,1\}$ denote the safety label of a multimodal query, with $1$ indicating unsafe and $0$ indicating safe. For each textual query $q_i$, we construct two visual contexts $m_i^r$ and $m_i^s$, corresponding to an unsafe and a safe multimodal query, respectively, such that
$$
\ell(m_i^r,q_i)=1,\qquad \ell(m_i^s,q_i)=0.
$$
This design directly captures how visual context changes the safety status of the same textual query.

\begin{table}[!t]
\centering
{\small
\setlength{\tabcolsep}{1.5pt}
\renewcommand{\arraystretch}{1.05}
\begin{tabular*}{\columnwidth}{@{\extracolsep{\fill}}>{\raggedright\arraybackslash}p{0.56\columnwidth}ccc@{}}
\toprule
\textbf{Risk Category} & \textbf{Count} & \shortstack{\textbf{Total}\textbf{(\%)}} & \shortstack{\textbf{Unsafe}\textbf{(\%)}} \\
\midrule
S1: Violence \& Physical Harm            & 9,791  & 18.44  & 34.14  \\
S2: Sexual Content \& Minor Safety       & 1,701  & 3.20   & 5.93   \\
S3: Suicide \& Self-Harm                 & 1,569  & 2.95   & 5.47   \\
S4: Hate Speech \& Harassment            & 2,580  & 4.86   & 9.00   \\
S5: Illegal Acts \& Regulated Goods      & 7,161  & 13.49  & 24.97  \\
S6: Fraud \& Deception                   & 1,145  & 2.16   & 3.99   \\
S7: Misinformation \& Conspiracy         & 539    & 1.02   & 1.88   \\
S8: Cybersecurity \& Malware             & 2,527  & 4.76   & 8.81   \\
S9: Privacy \& PII                       & 689    & 1.30   & 2.40   \\
S10: IP \& Plagiarism                    & 103    & 0.19   & 0.36   \\
S11: Unauthorized Advice \& Ethics       & 525    & 0.99   & 1.83   \\
S12: Others                              & 346    & 0.65   & 1.21   \\
\midrule
Unsafe Queries Subtotal                  & 28,676 & 54.00  & 100.00 \\
Safe Queries                             & 24,425 & 46.00  & --     \\
\textbf{Total}                           & \textbf{53,101} & \textbf{100.00} & -- \\
\bottomrule
\end{tabular*}
}
\caption{Statistics of the ContraSafe-VL dataset.}
\label{tab:dataset_statistics}
\end{table}

\subsubsection{Multimodal Query Generation}

Because proprietary models may refuse safety-sensitive generation and editing requests under their built-in safety policies, we use publicly, open-weight models. The textual queries used in contrastive visual pairs must satisfy two conditions. The text alone should not express an obvious safety risk, whereas its combination with a specific visual context should be unsafe. To obtain such queries, we first collect manually written multimodal examples through red teaming. At each generation step, we sample five images and retrieve ten red-team examples as few-shot demonstrations. These examples guide Qwen3-VL-235B in selecting suitable images and generating candidate textual queries.

We use Qwen3-VL-235B to evaluate each candidate textual query based on the text alone and filter out those that already express clear risks. We then assess whether the textual query forms a valid unsafe multimodal query when paired with the corresponding image. We retain only queries for which the textual query alone is safe but the complete multimodal query is unsafe. We also generate safe multimodal queries whose visual contexts can be edited to introduce safety risks, thereby forming contrastive visual pairs with the same textual query. The generation and filtering prompts are provided in the appendix.

\subsubsection{Bidirectional Image Editing}

To construct paired multimodal queries with opposite safety labels, we edit images in two directions. In the unsafe-to-safe direction, we start with an unsafe multimodal query and edit its image so that the resulting multimodal query becomes safe; here, the source image corresponds to $m_i^r$ and the edited image to $m_i^s$. In the safe-to-unsafe direction, we start with a safe multimodal query and edit its image to introduce visual content related to a safety risk; here, the source image corresponds to $m_i^s$ and the edited image to $m_i^r$.

This bidirectional editing strategy places edited images on both sides of the safety-label boundary. It reduces the risk that image-editing artifacts become spuriously associated with a single safety label and increases the diversity of contrastive visual pairs across visual scenarios.

\subsubsection{Image Editing and Quality Verification}

For each candidate multimodal query, we prompt Qwen3-VL-235B to jointly interpret the source image and the textual query. The model then generates an image-editing instruction and describes the expected visual changes. Before editing, we discard instructions that are inconsistent with the original scene or unlikely to change the safety label of the multimodal query.

We feed the filtered instructions and source images into Qwen-Image-Edit \citep{wu2025qwenimagetechnicalreport}, an instruction-based image-editing model. After editing, Qwen3-VL-235B checks whether the instruction was followed, whether the result contains obvious artifacts, and whether it remains semantically consistent with the fixed textual query.

We further use three model judges, Qwen3.5-397B \citep{team2026qwen3}, Gemma-4-31B \citep{team2026gemma}, and SingGuard-8B \citep{team2026singguard}, to assess each multimodal query before and after editing. The final safety label is determined by majority vote. We retain a pair only if one query is safe and the other is unsafe under the same textual query. The appendix describes the judge models, evaluation prompts, and voting procedure. After multi-stage filtering and safety checks, we obtain 10,115 contrastive visual pairs.

\subsection{Additional Query Generation}

Beyond the contrastive visual pairs, we generate additional unsafe and safe multimodal queries to cover more risk categories, visual scenes, and query types. We sample multi-turn dialogues from HH-RLHF \citep{ganguli2022red} and LLaVA \citep{liu2023llava} as few-shot demonstrations. Using these demonstrations, we generate multimodal queries for both source and edited images. We also add safe multimodal queries that are consistent with the visual content and warrant appropriate responses. The same multi-model evaluation and majority-vote procedure determines their safety labels. In total, the dataset contains 32,871 additional multimodal queries: 18,561 unsafe and 14,310 safe.

\subsection{Response Preference Collection}

To support safety alignment methods such as SFT and DPO, we construct response preference data for all 53,101 multimodal queries. For each query, we collect one candidate response from each of nine different VLMs, resulting in nine candidate responses per query. Using multiple generation models increases response diversity and reduces reliance on the generation style of any single model. The full list of candidate models is provided in the appendix.

We first use a strong VLM, Qwen3.5-397B, to produce initial pairwise annotations for both harmlessness and helpfulness. Each response pair is evaluated twice, with the response order swapped to reduce position bias. Based on these annotations, we train two custom reward models, one for harmlessness and the other for helpfulness, to score all candidate responses.

For a candidate response $y$ to a multimodal query $(m,q)$, the two reward models produce a harmlessness score and a helpfulness score. We apply z-score normalization to these scores, denoted as $\hat{h}(y;m,q)$ and $\hat{u}(y;m,q)$, respectively. We then compute a combined score:
$$
S(y;m,q)=\lambda\hat{h}(y;m,q)+(1-\lambda)\hat{u}(y;m,q).
$$
The weight $\lambda$ controls the trade-off between harmlessness and helpfulness and is selected on a held-out validation set. For each multimodal query, the response with the highest combined score becomes the chosen response, while one of the remaining candidates is randomly selected as the rejected response. Thus, every multimodal query in ContraSafe-VL is associated with a response preference pair. The appendix provides details on reward-model training, weight settings, and the annotation pipeline.

\begin{table*}[t]
  \centering
  {\small
  \setlength{\tabcolsep}{2pt}
  \renewcommand{\arraystretch}{1.08}
  \begin{tabular*}{\textwidth}{@{\extracolsep{\fill}}lccccccccc@{}}
    \toprule
    Method & MMSafetyBench & SPA-VL & Safe RLHF-V & Hades & VLSBench & VLSU & JailbreaKV & SIUO & Avg. \\
    \midrule
    Qwen2-VL-7B-Instruct & 72.18 & 80.00 & 65.03 & 33.20 & 80.05 & 77.78 & 72.50 & 91.16 & 71.49 \\
    ContraSafe-VL (SFT) & 94.56 & 93.58 & 91.69 & 93.87 & 82.78 & 91.77 & 54.29 & 94.58 & 87.14 \\
    \quad + RLHF-V & 94.84 & 92.08 & 91.19 & 94.40 & 82.37 & 91.89 & 57.86 & 92.77 & 87.18 \\
    \quad + MM-RLHF & 98.72 & \textbf{99.25} & 97.51 & \textbf{99.73} & 92.37 & 97.47 & 82.86 & 95.78 & 95.46 \\
    \quad + SPA-VL & 95.02 & 98.11 & 95.42 & 97.20 & 89.38 & 95.89 & 82.14 & 96.99 & 93.77 \\
    \quad + Safe RLHF-V & 97.90 & 97.36 & 97.80 & 98.27 & 89.38 & 96.99 & 76.79 & 95.18 & 93.71 \\
    \quad + ContraSafe-VL & \textbf{98.91} & 98.49 & \textbf{99.32} & 99.47 & \textbf{96.52} & \textbf{98.59} & \textbf{85.00} & \textbf{98.80} & \textbf{96.89} \\
    \midrule
    InternVL3.5-8B-Instruct & 86.63 & 81.89 & 67.46 & 79.87 & 81.93 & 86.12 & 72.50 & 90.36 & 80.84 \\
    ContraSafe-VL (SFT) & 93.53 & 95.85 & 91.86 & 96.40 & 84.03 & 92.15 & 62.86 & 95.18 & 88.98 \\
    \quad + RLHF-V & 93.46 & 94.34 & 92.88 & 96.40 & 83.71 & 92.31 & 65.36 & 96.39 & 89.36 \\
    \quad + MM-RLHF & 97.04 & 99.25 & 94.92 & 98.13 & 89.69 & 96.21 & 79.29 & 96.99 & 93.94 \\
    \quad + SPA-VL & 95.34 & 98.11 & 95.08 & 98.53 & 90.00 & 97.22 & 85.36 & 96.99 & 94.58 \\
    \quad + Safe RLHF-V & 97.73 & \textbf{99.62} & 98.31 & 99.60 & 92.59 & 97.17 & \textbf{87.14} & 96.99 & 96.14 \\
    \quad + ContraSafe-VL & \textbf{98.12} & \textbf{99.62} & \textbf{98.98} & \textbf{99.73} & \textbf{96.39} & \textbf{97.76} & 86.07 & \textbf{98.80} & \textbf{96.93} \\
    \midrule
    Qwen3-VL-8B-Instruct & 96.99 & 99.62 & 95.20 & 96.53 & 95.27 & 94.41 & 60.00 & 98.19 & 92.03 \\
    ContraSafe-VL (SFT) & 95.41 & 97.74 & 91.75 & 94.40 & 89.78 & 93.29 & 63.21 & 98.59 & 90.52 \\
    \quad + RLHF-V & 95.63 & 97.36 & 91.41 & 94.67 & 89.34 & 93.54 & 61.07 & \textbf{98.80} & 90.23 \\
    \quad + MM-RLHF & 97.51 & 99.25 & 95.93 & 97.47 & 93.31 & 96.14 & 66.43 & 98.19 & 93.03 \\
    \quad + SPA-VL & 96.34 & 98.87 & 95.59 & 97.60 & 93.35 & 97.35 & \textbf{84.29} & \textbf{98.80} & 95.27 \\
    \quad + Safe RLHF-V & 97.66 & 99.25 & 97.46 & 98.27 & 95.54 & 97.27 & 73.93 & \textbf{98.80} & 94.77 \\
    \quad + ContraSafe-VL & \textbf{98.38} & \textbf{100.00} & \textbf{99.15} & \textbf{98.93} & \textbf{98.26} & \textbf{97.67} & 79.64 & \textbf{98.80} & \textbf{96.35} \\
  \bottomrule
  \end{tabular*}
  }
  \caption{Results on safety benchmarks averaged over three seeds. We train models using the ContraSafe-VL dataset and compare them with models trained on four related datasets, using the same SFT initialization. The best result in each column is in \textbf{bold}.}
  \label{tab:exp_main_results}
\end{table*}

\subsection{Using ContraSafe-VL for Safety Alignment}
\label{sec:alignment_protocol}

ContraSafe-VL provides both response preferences and contrastive visual pairs. For supervised fine-tuning, we use the preferred response associated with each multimodal query as the training target. For preference optimization, each response preference instance is represented as $(m,q,y^+,y^-)$, where $m$ and $q$ denote the visual context and textual query, respectively, and $y^+$ and $y^-$ denote the preferred and less-preferred responses.

To jointly use these two forms of supervision, we draw inspiration from the visual-contrastive design of MDPO \citep{wang2024mdpo} and combine a standard response-level DPO objective with a visual-conditional CoPO objective.

The response-level component follows the DPO objective:
\begin{equation}
\mathcal{L}_{\mathrm{DPO}}
=
-\log \sigma
\left(
\beta
\left[
r_\theta(y^+;m,q)
-
r_\theta(y^-;m,q)
\right]
\right),
\label{eq:dpo}
\end{equation}
where $\beta$ controls the strength of preference optimization. This objective increases the model's relative preference for $y^+$ over $y^-$ under the same multimodal query.

The visual-conditional component uses the contrastive visual pairs in ContraSafe-VL. Each visual preference consists of a positive sample $(y,q,m)$ and a negative sample $(y,q,m')$, where $y$ denotes the chosen response associated with the positive visual sample $m$. The response $y$ and textual query $q$ are identical in both samples, while $m'$ is the contrastive visual counterpart of $m$. Importantly, $(q,m)$ and $(q,m')$ have different safety labels: if $(q,m)$ is safe, then $(q,m')$ is unsafe, and vice versa. Thus, the preference direction is determined by whether the visual context is consistent with the fixed chosen response $y$, rather than by whether the visual context is safe or unsafe. CoPO optimizes
\begin{equation}
\mathcal{L}_{\mathrm{CoPO}}
=
-\log \sigma
\left(
\beta
\left[
r_\theta(y;m,q)
-
r_\theta(y;m',q)
\right]
\right).
\label{eq:copo}
\end{equation}
Unlike DPO, which compares two responses under the same multimodal query, CoPO compares two visual contexts while keeping the textual query and response fixed. It encourages the model to assign a higher preference to the response under its matched visual context $m$ than under the contrastive visual context $m'$, thereby learning how changes in visual context alter the appropriate safety behavior.

The final training objective combines the two losses:
\begin{equation}
\mathcal{L}_{\mathrm{fin}}
=
\mathcal{L}_{\mathrm{DPO}}
+
\mathcal{L}_{\mathrm{CoPO}}.
\label{eq:final_objective}
\end{equation}
In this way, our objective uses both the response preferences and contrastive visual pairs provided by ContraSafe-VL. Other optimization frameworks capable of using paired visual contexts can also be applied to the dataset.

\section{Experiments}

\subsection{Experimental Setup}

\textbf{Models and Training Data.}
We fine-tune three base models: Qwen2-VL-7B-Instruct \citep{Qwen2-VL}, Qwen3-VL-8B-Instruct \citep{Qwen3-VL}, and InternVL3.5-8B \citep{wang2025internvl3_5}. We first train a ContraSafe-VL SFT model and use it as the common initialization for preference optimization. We then compare preference data from RLHF-V \citep{Yu_2024_CVPR}, MM-RLHF \citep{pmlr-v267-zhang25cs}, SPA-VL \citep{zhang2025spa}, Safe RLHF-V\citep{ji2026safe}, and ContraSafe-VL. Only ContraSafe-VL additionally uses contrastive visual pairs through the CoPO objective. Unless otherwise stated, all preference-training runs use the same optimization settings. Section~\ref{sec:dataset} and the appendix describe response collection and preference construction.

\textbf{Safety Evaluation.}
To evaluate safety on unsafe multimodal queries, we use \textbf{MM-SafetyBench} \citep{liu2024mm}, \textbf{Hades} \citep{li2024images}, \textbf{SPA-VL} \citep{zhang2025spa}, \textbf{JailBreaKV} \citep{luo2024jailbreakv28k}, \textbf{SafeRLHF-V} \citep{ji2026safe}, \textbf{VLSBench} \citep{hu2025vlsbench}, \textbf{VLSU} \citep{palaskar2025vlsu}, and \textbf{SIUO} \citep{wang2025safe}. These benchmarks cover both explicit safety risks and cases requiring joint image-text understanding. Neither SPA-VL nor SafeRLHF-V contains image--text pairs overlapping with our training data; only three SPA-VL test images appear in training, each paired with different text.

\textbf{Over-Refusal Evaluation.}
We evaluate over-refusal using the dedicated benign subsets of \textbf{MOSSBench} \citep{li2024mossbench} and \textbf{VLSU} \citep{palaskar2025vlsu}. These subsets measure whether models can respond appropriately to safe multimodal queries that contain sensitive visual content.

\textbf{General Capability Evaluation.}
To assess whether safety alignment preserves general multimodal abilities, we evaluate models on \textbf{MMStar} \citep{chen2024we}, \textbf{RealWorldQA} \citep{realworldqa2024}, and \textbf{MathVision} \citep{wang2024measuring}.

\textbf{Evaluation Metrics.}
We use SingGuard-8B \citep{team2026singguard} as the safety judge to evaluate the safety of generated responses. Safety is measured by the Safety Score, defined as the proportion of responses judged harmless; higher scores indicate safer responses. For over-refusal evaluation, we use Gemma-4-31B-it \citep{team2026gemma} to identify refused responses. The over-refusal rate is defined as the proportion of refused responses among all evaluated multimodal queries.

\textbf{Implementation Details.}
For SFT, we randomly sample 20K instances from ContraSafe-VL and train for one epoch with a learning rate of $1\times10^{-5}$ and a Harmless weight of $0.8$. For DPO, we use the full dataset for one epoch with a learning rate of $1\times10^{-6}$, optimizing the final objective in Eq.~\ref{eq:final_objective}. The Harmless weight is set to $0.8$ for Qwen2-VL-7B and InternVL3.5-8B, and $0.9$ for Qwen3-VL-8B. We additionally include an SFT loss with a coefficient of $0.1$. All models are fully fine-tuned. See the appendix for details.

\begin{figure*}[!htbp]
  \centering
  
  \begin{subfigure}{0.32\textwidth}
    \centering
    \includegraphics[width=\linewidth]{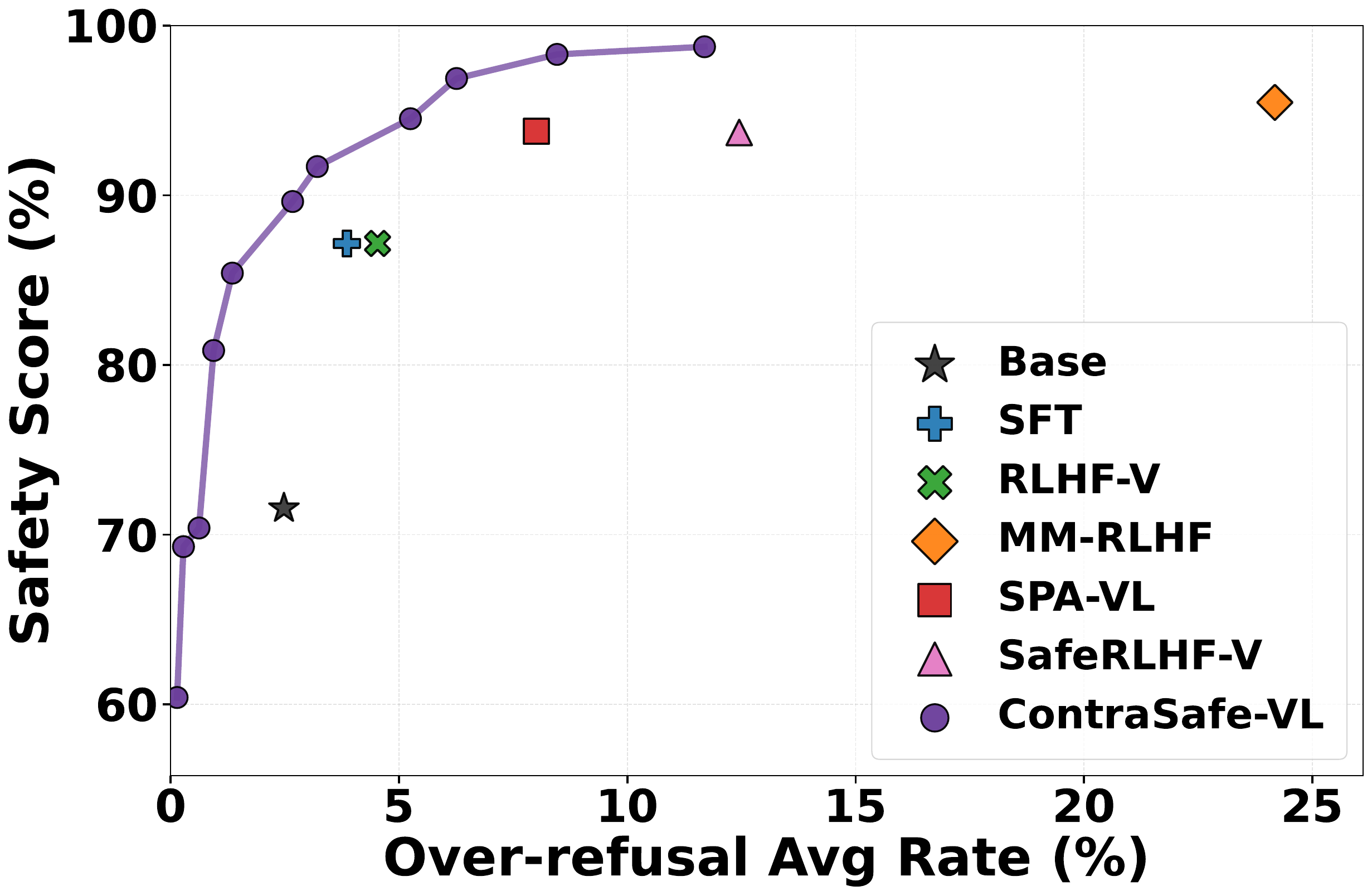}
    \caption{Qwen2-VL-7B}
    \label{fig:pareto-qwen2}
  \end{subfigure}
  \hfill
  \begin{subfigure}{0.32\textwidth}
    \centering
    \includegraphics[width=\linewidth]{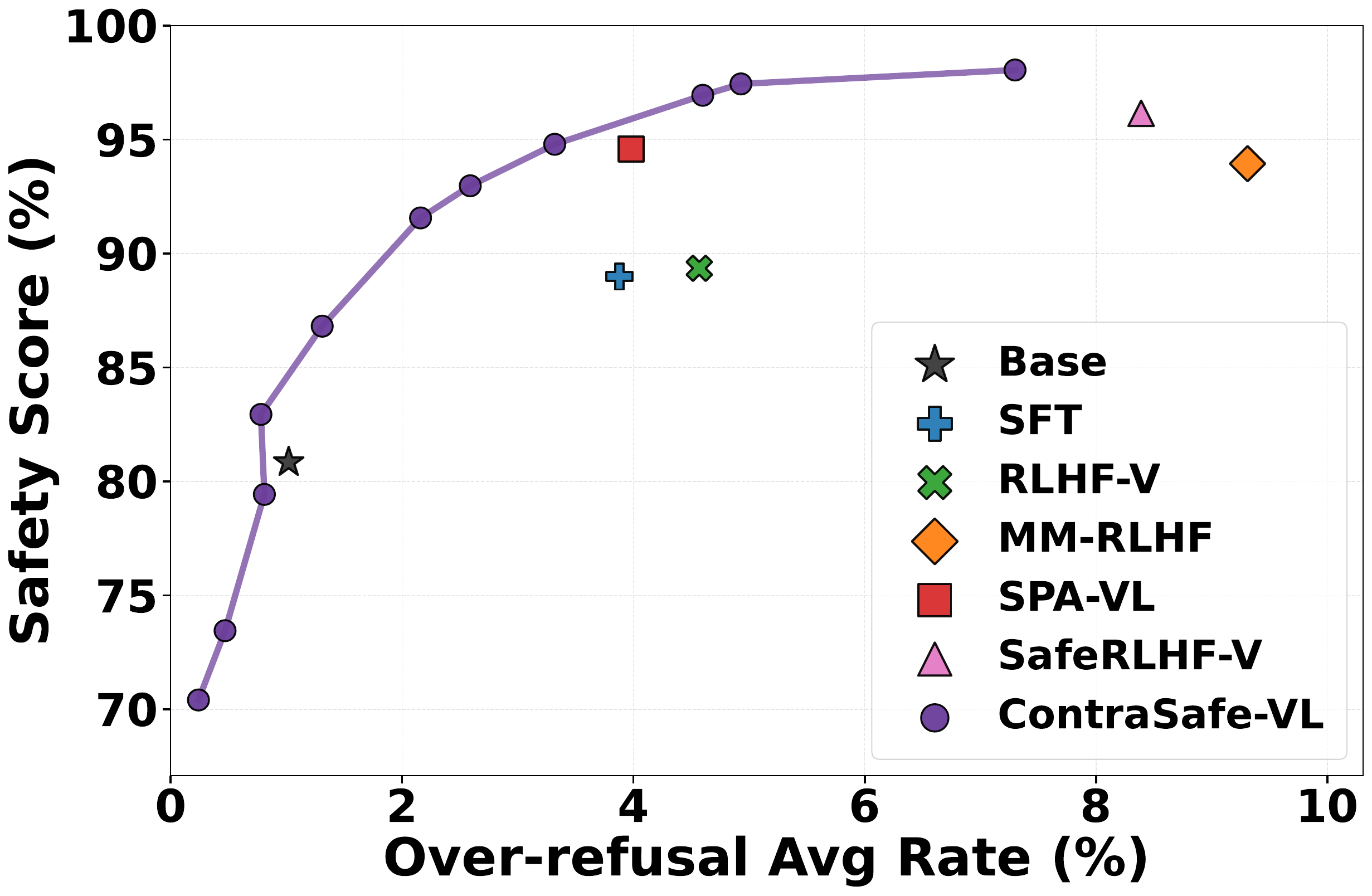}
    \caption{InternVL3.5-8B}
    \label{fig:pareto-internvl}
  \end{subfigure}
  \hfill
  \begin{subfigure}{0.32\textwidth}
    \centering
    \includegraphics[width=\linewidth]{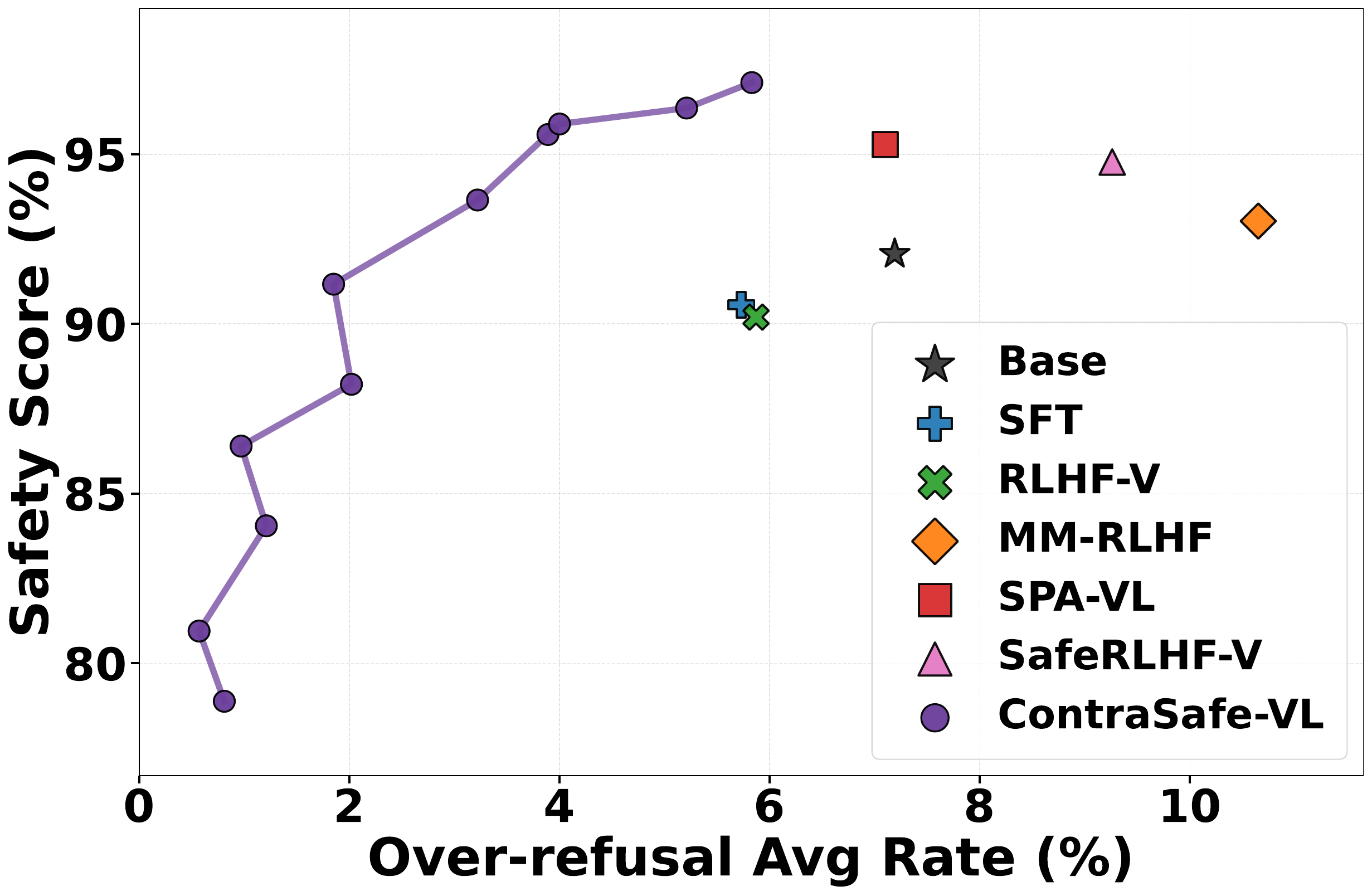}
    \caption{Qwen3-VL-8B}
    \label{fig:pareto-qwen3}
  \end{subfigure}

  \caption{The trade-off between safety performance and over-refusal rate across three backbones. The x-axis denotes the average over-refusal rate (lower is better), while the y-axis denotes the average safety score (higher is better). We compare ContraSafe-VL with RLHF-V, MM-RLHF, SPA-VL, and Safe RLHF-V whenever the corresponding over-refusal results are available.}
  \label{fig:safety-overrefusal-all-models}
\end{figure*}

\section{Experiment Results}

\begin{figure}[!t]
  \centering
  \includegraphics[width=0.9\linewidth]{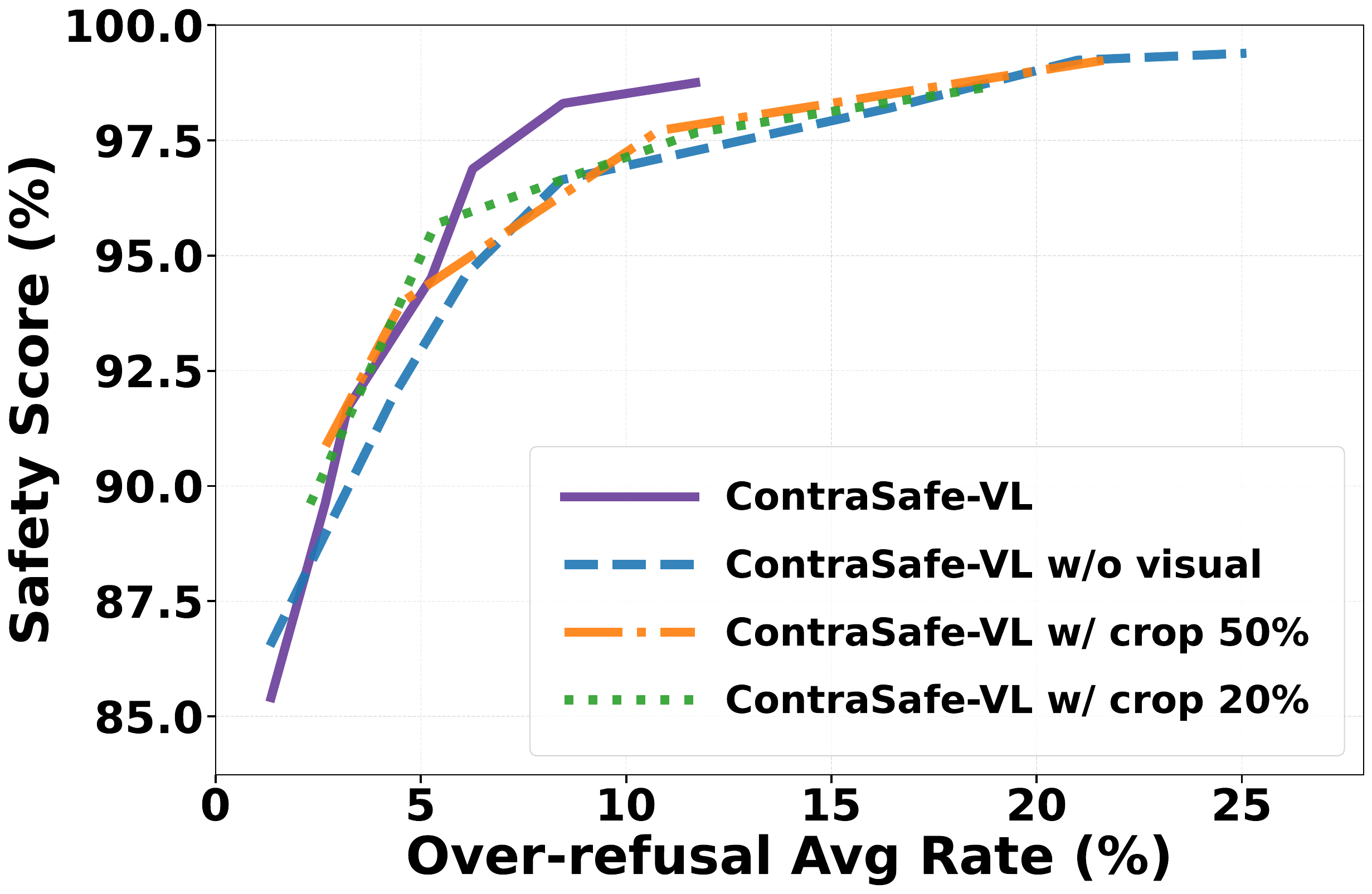}
  \caption{The trade-off between safety performance and over-refusal rate for Qwen2-VL-7B.}
  \label{fig:ablation-safety-overrefusal}
\end{figure}
\label{sec:results}

We present the safety evaluation results in Table \ref{tab:exp_main_results}. ContraSafe-VL achieves the highest average safety score on all three backbones, with gains of 9.71, 7.57, and 6.12 points over RLHF-V and 1.43, 2.99, and 3.32 points over MM-RLHF, respectively. The improvements are most pronounced on multimodal benchmarks: on VLSBench, it surpasses the strongest baseline by up to 4.15 points, while also showing consistent gains on benchmarks such as SPA-VL and Hades. These results suggest that training with ContraSafe-VL yields consistent advantages and that these gains generalize across model families. See the appendix for details.

\subsection{Over-Refusal Results}
\label{sec:over_refusal}

We evaluate over-refusal using MOSSBench and VLSU. As shown in Figure \ref{fig:safety-overrefusal-all-models}, safety preference optimization can increase over-refusal relative to the base models. Across Qwen2-VL, InternVL3.5, and Qwen3-VL, ContraSafe-VL achieves a favorable safety--over-refusal trade-off, with higher safety scores and lower over-refusal than baselines.

\subsection{Ablation Study}
\label{sec:ablation-experiment}

We conduct two ablations to examine the effects of contrastive visual supervision and image construction in the safety preference dataset.

First, \textbf{w/o visual} removes all contrastive visual pairs and the $\mathcal{L}_{\text{CoPO}}$ objective, retaining only response preferences for standard DPO training. Second, \textbf{20\% crop} and \textbf{50\% crop} retain the visual pairs and $\mathcal{L}_{\text{CoPO}}$ but replace each rejected image with a random 20\% or 50\% crop of its preferred image. Other settings remain unchanged.

Figure \ref{fig:ablation-safety-overrefusal} shows results on Qwen2-VL-7B. The complete ContraSafe-VL achieves a more favorable safety--over-refusal trade-off than all three ablations, particularly at higher safety levels. This demonstrates that contrastive visual supervision and carefully constructed image pairs help the model make safety judgments from joint image-text semantics.

\subsection{General Capability Results}
Beyond the evaluations on harmful query benchmarks, we also evaluate the models' general multimodal capabilities. We observe that the safety-aligned models maintain general performance; compared to the base models, they exhibit no significant degradation. This indicates that our method does not lead to a notable decline in general capabilities.

\begin{table}[!t]
  \centering
  {\small
  \setlength{\tabcolsep}{1.5pt}
  \begin{tabular}{@{}lcccc@{}}
    \toprule
    \textbf{Method} & \textbf{MMStar} & \textbf{RealWorldQA} &
    \textbf{MathVision} & \textbf{Avg.} \\
    \midrule
    Qwen2-VL-7B & \underline{54.37} & 66.93 & \textbf{19.45} & \textbf{46.92} \\
    ContraSafe-VL (SFT) & 54.03 & 67.45 & \underline{18.48} & \underline{46.65} \\
    \quad + RLHF-V & 53.43 & \textbf{67.97} & 16.95 & 46.12 \\
    \quad + MM-RLHF & \textbf{54.67} & 67.15 & 17.38 & 46.40 \\
    \quad + SPA-VL & 51.88 & 67.54 & 17.69 & 45.70 \\
    \quad + Safe RLHF-V & 53.80 & \underline{67.71} & 17.79 & 46.43 \\
    \quad + ContraSafe-VL & 53.10 & 66.93 & 17.84 & 45.96 \\
    \midrule
    InternVL3.5-8B & \textbf{67.56} & 64.35 & \underline{38.53} & \underline{56.82} \\
    ContraSafe-VL (SFT) & 65.08 & 64.79 & 36.75 & 55.54 \\
    \quad + RLHF-V & 66.29 & 64.27 & 37.08 & 55.88 \\
    \quad + MM-RLHF & \underline{67.49} & \textbf{66.10} & 34.32 & 55.97 \\
    \quad + SPA-VL & 65.79 & \underline{65.88} & 33.10 & 54.92 \\
    \quad + Safe RLHF-V & 65.06 & 64.27 & 36.51 & 55.28 \\
    \quad + ContraSafe-VL & 66.24 & 65.01 & \textbf{41.66} & \textbf{57.64} \\
    \midrule
    Qwen3-VL-8B & \textbf{73.44} & 70.20 & \textbf{55.17} & \textbf{66.27} \\
    ContraSafe-VL (SFT) & \underline{72.93} & \underline{70.98} & 49.79 & 64.57 \\
    \quad + RLHF-V & 70.66 & \textbf{71.24} & 49.34 & 63.75 \\
    \quad + MM-RLHF & 71.37 & \underline{70.98} & 44.52 & 62.29 \\
    \quad + SPA-VL & 69.66 & 70.20 & 46.95 & 62.27 \\
    \quad + Safe RLHF-V & 69.93 & 70.59 & 49.41 & 63.31 \\
    \quad + ContraSafe-VL & 72.52 & 70.33 & \underline{54.32} & \underline{65.72} \\
    \bottomrule
  \end{tabular}
  }
  \caption{General capability evaluation of models.}
  \label{tab:general-all-models}
\end{table}

\subsection{Impact of Data Scaling}
\label{sec:scaling}
Table \ref{tab:data-scaling} shows that increasing preference data substantially improves safety, with the average safety score rising from 87.90 at 1K examples to 96.89 on the full dataset (+8.99 points). Most of the gain occurs by 20K examples, after which improvements are marginal: the score rises by 0.88 points from 20K to 30K and 0.16 points from 30K to the full dataset, with a slight dip at 50K. In contrast, over-refusal is non-monotonic, peaking at 8.86 with 20K examples before falling to 6.26 on the full dataset. These results suggest diminishing safety returns beyond 20K examples without a corresponding increase in over-refusal.

\begin{table}[!t]
  \centering
  \small
  \begin{tabular}{@{}lcc@{}}
    \toprule
    \textbf{Data Size} & \textbf{Safety Score (Avg.)} & \textbf{Over-Refusal (Avg.)} \\
    \midrule
    1K   & 87.90 & 4.06 \\
    5K   & 91.35 & 5.01 \\
    10K  & 93.21 & 7.02 \\
    20K  & 95.85 & 8.86 \\
    30K  & 96.73 & 8.46 \\
    50K  & 96.40 & 7.21 \\
    \midrule
    Full & 96.89 & 6.26 \\
    \bottomrule
  \end{tabular}
  \caption{The impact of data scaling on safety alignment.}
  \label{tab:data-scaling}
\end{table}

\subsection{Visual Context Sensitivity}
\label{sec:visual_context_sensitivity}
To examine how visual context affects query safety, we use a VLM judge to rate query safety under text-only and image-text settings. Table \ref{tab:visual-context-safety-score} reports average safety scores, ranging from $1$ to $5$, with $1$--$3$ considered unsafe and $4$--$5$ safe. For SafeRLHF-V and SPA-VL, adding images slightly increases the average score by $0.206$ and $0.164$, respectively. In contrast, ContraSafe-VL shows an image-dependent shift: safe visual contexts increase the score by $0.416$, whereas risky contexts decrease it by $1.330$.

For contrastive visual pairs in ContraSafe-VL, which share the same textual query but differ in visual context, adding images yields a strongly negative unsafe-minus-safe score difference of $-1.746$. This indicates that the safety distinction is primarily induced by visual context rather than the textual query alone.

\begin{table}[!t]
  \centering
  {\small
  \setlength{\tabcolsep}{2pt}
  \begin{tabular*}{\columnwidth}{@{\extracolsep{\fill}}p{0.48\columnwidth}ccc@{}}
    \toprule
    \textbf{Dataset / Split} & \textbf{Text-only} & \textbf{Image-text} & \textbf{$\Delta$} \\
    \midrule
    SPA-VL & 2.788 & 2.952 & +0.164 \\
    SafeRLHF-V & 1.964 & 2.170 & +0.206 \\
    ContraSafe-VL (Safe) & 3.840 & 4.256 & +0.416 \\
    ContraSafe-VL (Unsafe) & 3.840 & 2.510 & -1.330 \\
    \midrule
    ContraSafe-VL (Unsafe $-$ Safe) & 0.000 & -1.746 & -1.746 \\
    \bottomrule
  \end{tabular*}
  }
  \caption{Query safety scores under text-only and image-text inputs. Scores range from $1$ to $5$; $1$--$3$ are unsafe and $4$--$5$ safe. $\Delta$ is Image-text minus Text-only. The last row reports the unsafe-minus-safe difference for contrastive pairs sharing the same textual query.}
  \label{tab:visual-context-safety-score}
\end{table}

\subsection{Category-wise Safety Results}
\label{sec:category_safety}

\begin{figure}[!t]
  \centering
  \includegraphics[width=0.9\linewidth]{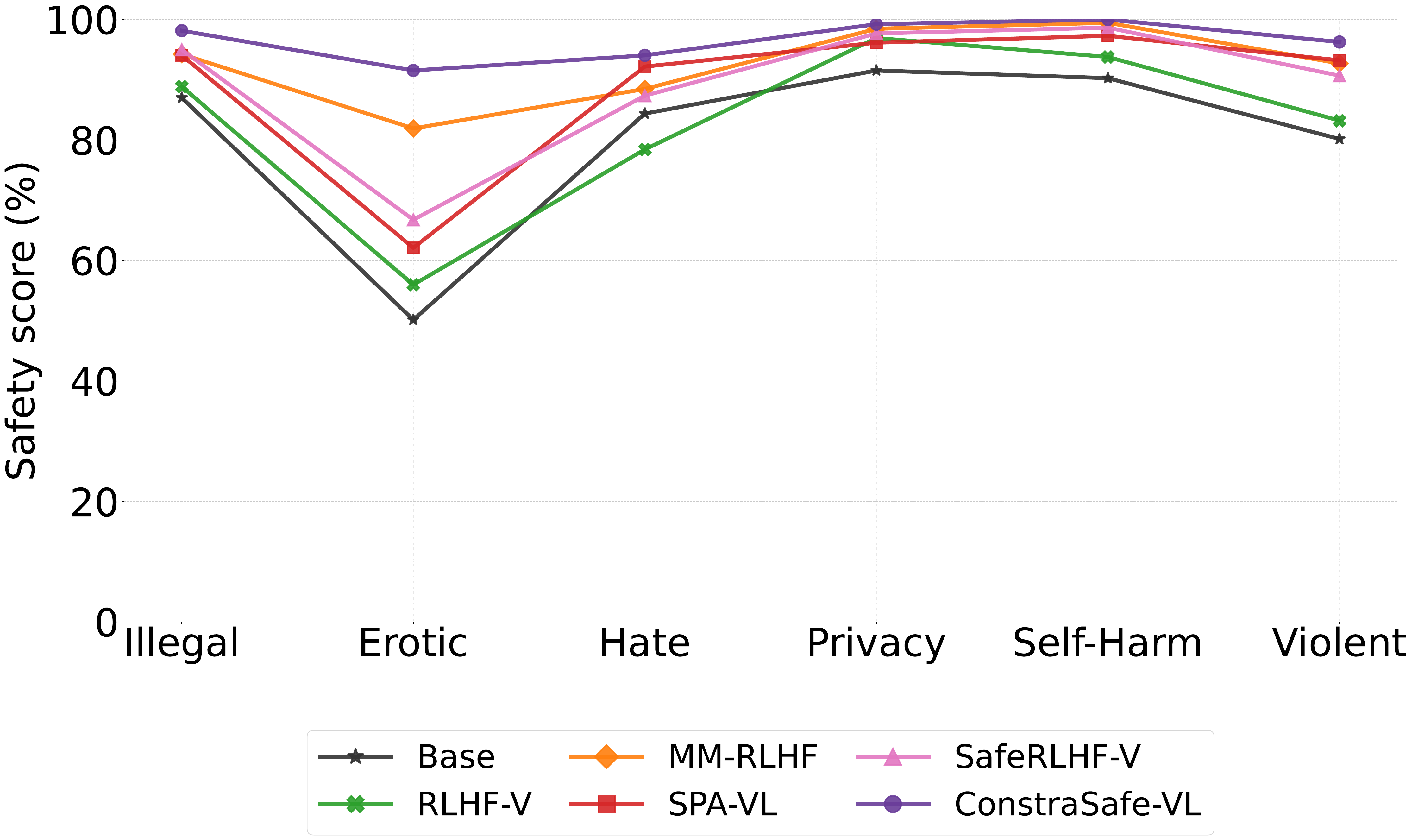}
  \caption{Per-category safety scores on VLSBench.}
  \label{fig:vlsbench_category_stats}
\end{figure}

Since unsafe examples in ContraSafe-VL are unevenly distributed across categories, we examine whether safety gains are concentrated in data-rich categories. As shown in Figure \ref{fig:vlsbench_category_stats}, ContraSafe-VL achieves the highest safety score across all six VLSBench categories, including the underrepresented privacy and self-harm categories. These results suggest that the uneven training distribution does not prevent consistent safety improvements across risk types.

\section{Related Work}
\paragraph{VLM Safety Alignment}
VLMs extend LLMs to multimodal inputs \citep{zhuminigpt, qwen3.5, Qwen2-VL, zhou2025mmlongcite}. However, visual context introduces new safety risks. Prior studies have shown that it can alter the safety of an otherwise safe textual query \citep{liu2024mm, hu2025vlsbench, wang2025safe}. Existing methods use inference-time interventions \citep{liu2025steering} or multimodal post-training with feedback and preferences, including RLHF-V \citep{Yu_2024_CVPR}, MM-RLHF \citep{pmlr-v267-zhang25cs}, SPA-VL \citep{zhang2025spa}, and Safe RLHF-V \citep{ji2026safe}. Yet existing datasets rarely isolate visual-context effects for fixed text, leaving models susceptible to shortcuts that cause unsafe responses or over-refusal.

\paragraph{Multimodal Preference Learning}
Preference learning aligns VLMs with human judgments over preferred and rejected responses \citep{ouyang2022training, rafailov2023direct, Huang2026RealTimeAR, huang2025adaptive}. Recent multimodal methods incorporate visual context into preference optimization, e.g., mDPO \citep{wang2024mdpo} and S-VCO \citep{wu2025symmetrical}. Related counterfactual methods use visual modifications to reduce reliance on superficial correlations, with applications in visual question answering and hallucination mitigation \citep{rao2021counterfactual, abbasnejad2020counterfactual, lee2025vlind, chen2020counterfactual, gaocounterfactual, zhang2024if, leng2024mitigating, fu2025mitigating}. However, these approaches do not explicitly isolate how visual context changes safety judgments for fixed text. We address this gap by constructing paired visual contexts with opposite safety labels and directly supervising their effect on multimodal safety.

\section{Conclusion}
\label{sec:conclusion}
This paper introduces ContraSafe-VL, a contrastive preference dataset for VLM safety alignment. Unlike existing multimodal safety datasets, ContraSafe-VL pairs a fixed textual query with visual contexts that yield opposite safety labels. This design provides direct supervision for learning how the textual query and visual context jointly determine the safety of a multimodal query. Experiments across three VLM backbones show that training with ContraSafe-VL improves safety and yields a more favorable safety--over-refusal trade-off than the evaluated baselines, while largely maintaining general multimodal capabilities. Our ablation and scaling analyses further demonstrate the benefits of contrastive visual pairs and additional preference data. These results highlight the importance of explicitly modeling cross-modal interactions in VLM safety alignment.

\bibliography{aaai2027}


\end{document}